\documentclass[11pt,a4paper]{article}
\usepackage[nohyperref]{emnlp}

\usepackage{times}
\usepackage{latexsym}
\usepackage{amssymb}
\usepackage{url}
\usepackage{graphicx}
\usepackage{amsmath}
\usepackage{breqn}
\usepackage{algorithm}
\usepackage{algorithmicx}  
\usepackage{algpseudocode} 
\usepackage{url}
\usepackage{xcolor}
\usepackage{multirow}
\usepackage{subfigure}
\usepackage{color}
\usepackage{enumitem}
\usepackage{booktabs}

\aclfinalcopy 

\newcommand{\comments}[1]{}

\title{Dialog State Tracking with Reinforced Data Augmentation}

\author{Yichun Yin, Lifeng Shang, Xin Jiang, Xiao Chen, Qun Liu\\
Huawei Noah's Ark Lab\\
{\{yinyichun, shang.lifeng, jiang.xin, chen.xiao2, qun.liu\}@huawei.com}}

\date{}

\begin{document}
\maketitle

\begin{abstract}

Neural dialog state trackers are generally limited due to the lack of quantity and diversity of annotated training data. In this paper, we address this difficulty by proposing a reinforcement learning (RL) based framework for data augmentation that can generate high-quality data to improve the neural state tracker. Specifically, we introduce a novel {\it contextual bandit generator} to learn fine-grained augmentation policies that can generate new effective instances by choosing suitable replacements for specific context. Moreover, by alternately learning between the {\it generator} and the state {\it tracker}, we can keep refining the generative policies to generate more high-quality training data for neural state tracker. Experimental results on the WoZ and MultiWoZ (restaurant) datasets demonstrate that the proposed framework significantly improves the performance over the state-of-the-art models, especially with limited training data.
\end{abstract}

\vspace{-15pt}
\section{Introduction}
\vspace{-5pt}
\label{introduction}



With the increasing popularity of intelligent assistants such as Alexa, Siri and Google Duplex, the research on spoken dialog systems has gained a great deal of attention in recent years~\cite{gao2018neural}. Dialog state tracking (DST)~\cite{williams2013dialog} is an essential component of most spoken dialog systems, aiming to track user's goal at each step in a dialog. Based on that, the dialog agent decides how to converse with the user. In a \textit{slot-based} dialog system, the dialogue states are typically formulated as a set of slot-value pairs and one concrete example is as follows: 

\begin{table}[h]
\centering 
\begin{small}
\begin{tabular}{|p{7.3cm}}
{\fontsize{9.8}{7.2}\selectfont  \textbf{User:} Grandma wants Italian, any suggestions?}\\
\texttt{State: inform(food=\underline{Italian})}\\  [0.5ex]  
{\fontsize{9.8}{7.2}\selectfont \textbf{Agent:} Would you prefer south or center?}\\  [0.5ex] 
{\fontsize{9.8}{7.2}\selectfont\textbf{User:} It doesn't matter. Whichever is less expensive. }\\
\texttt{State: inform(food=\underline{Italian}, price=cheap, area=\underline{don't care})}\\ [0.5ex] 
\end{tabular}  
\end{small}
\end{table}

\begin{figure}[t]
\centering
\includegraphics[width=0.875\linewidth]{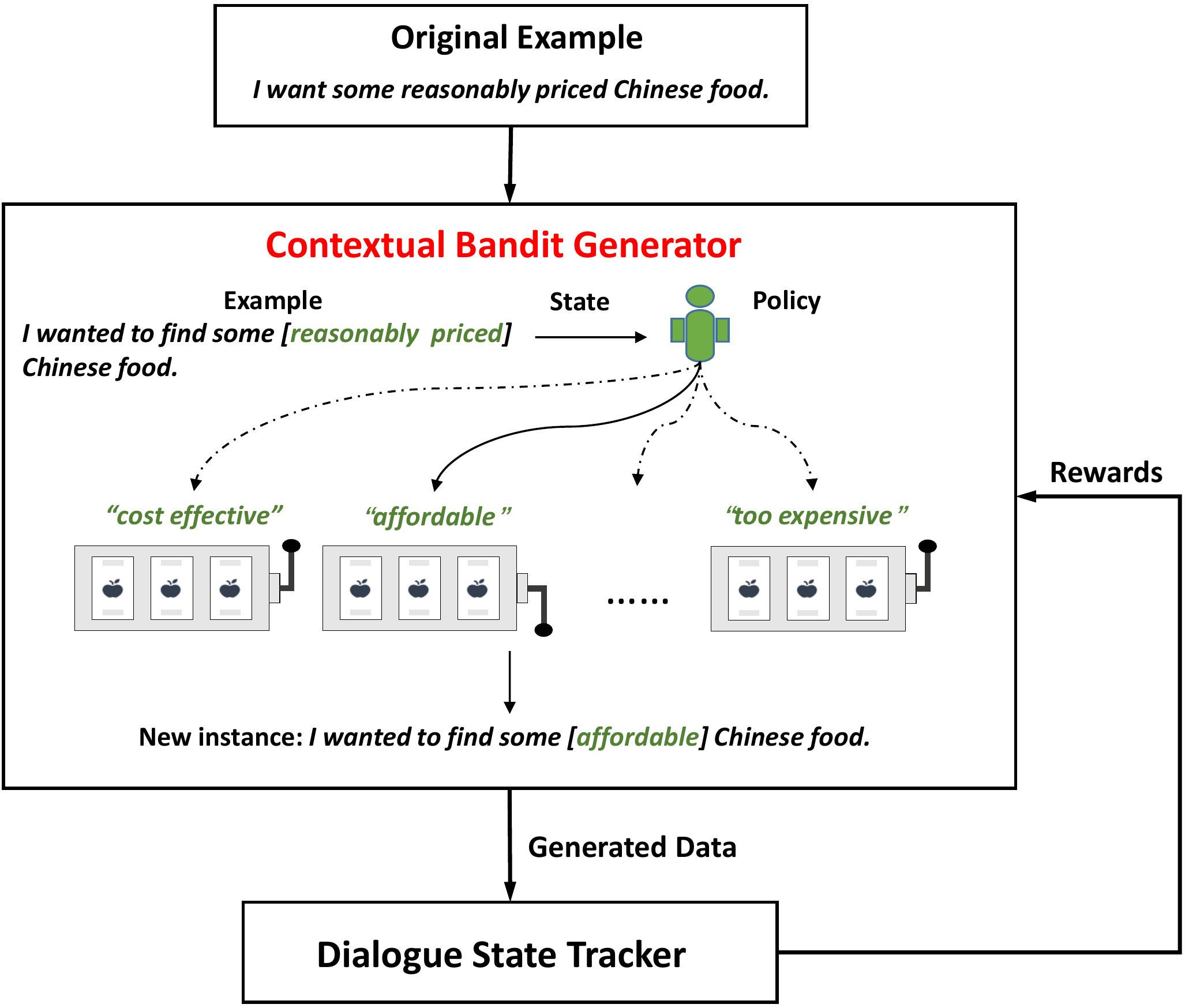}
\caption{An overview of our framework. Given a dataset, we induce new instances using the RL-based Generator to improve the DST Tracker. The Generator is trained with the rewards from the Tracker. The learning process is performed in an alternate manner.} 
\label{fig:rda}
\vspace{-15pt}
\end{figure}

The state-of-the-art models for DST are based on neural network~\cite{henderson2014word,mrkvsic2017neural,zhong2018global,D18-1299,sharma2019improving}. They typically predict the probabilities of the candidate slot-value pairs with the user utterance, previous system actions or other external information as inputs, and then determine the final value of each slot based on the probabilities. Although the neural network based methods are promising with advanced deep learning techniques such as gating and self-attention mechanisms~\cite{lin2017structured,vaswani2017attention}, the data-hungry nature makes them difficult to generalize well to the scenarios with limited or sparse training data. 

To alleviate the data sparsity in DST, we propose a reinforced data augmentation ({RDA}) framework to increase both the amount and diversity of the training data. The {RDA} learns to generate high-quality labeled instances, which are used to re-train the neural state trackers to achieve better performances. As shown in Figure~\ref{fig:rda}, the {RDA} consists of two primary modules: {Generator} and {Tracker}. The two learnable modules alternately learn from each other during the training process. On one hand, the {Generator} module is responsible for generating new instances based on a parameterized generative policy, which is trained with the rewards from the {Tracker} module. The {Tracker}, on the other hand, is refined via the newly generated instances from the {Generator}. 

Data augmentation performs perturbation on the original dataset without actually collecting new data, which has been widely used in the field of computer vision~\cite{krizhevsky2012imagenet,cubuk2018autoaugment} and speech recognition~\cite{ko2015audio}, but relatively limited in natural language processing~\cite{kobayashi2018contextual}. The reason is that, in contrast to image augmentation (e.g., rotating or flipping images), it is significantly more difficult to augment text because it requires preserving the semantics and fluency of newly augmented data. In this paper, to derive a more general and effective policy for text data augmentation, we adopt a coarse-to-fine strategy to model the generation process. Specifically, we initially use some coarse-grained methods to get candidates (such as {\it cost effective}, {\it affordable} and {\it too expensive} in Figure~\ref{fig:rda}), some of which are inevitably noisy or unreliable for the specific sentence context. We then adopt RL to learn the policies for selecting high quality candidates to generate new instances, where the total rewards are obtained from the {Tracker}. After learning the {Generator}, we use it to induce more training data to re-train the {Tracker}. Accordingly, the {Tracker} will further provide more reliable rewards to the~{Generator}. With alternate learning, we can progressively improve the generative policies for data augmentation and at the same time learn the better {Tracker} with the augmented data.

To demonstrate the effectiveness of the proposed {RDA} framework in DST, we conduct extensive experiments with the WoZ~\cite{wen2017network} and MultiWoZ~(restaurant)~\cite{budzianowski2018multiwoz} datasets. The results show that our model consistently outperforms the strong baselines and achieves new state-of-the-art results. In addition, the effects of the hyper-parameter choice on performance are analyzed and case studies on the policy network are performed. 

The main contributions of this paper include:

\vspace{-2mm}
\begin{itemize}
\setlength\itemsep{-0.2em}
\item We propose a novel framework of data augmentation for dialog state tracking, which can generate high-quality labeled data to improve neural state trackers.
\item We use RL for the {Generator} to produce effective text augmentation.
\item We demonstrate the effectiveness of the proposed framework on two datasets, showing that the {RDA} can consistently boost the state-tracking performance and obtain new state-of-the-art results.
\end{itemize}

\comments{In the remainder of this paper, we start by presenting the framework of {RDA} in Section 2. Then we report the experimental results in Section 3. After that we propose the related work in Section 4. In Section 5 we conclude the paper.}

\section{Reinforced Data Augmentation}
We elaborate on our framework in three parts: the Tracker module, the Generator module, and the alternate learning algorithm.
\subsection{Tracker Module}
\begin{figure}[t]
\begin{center}
\includegraphics[width=0.95\linewidth]{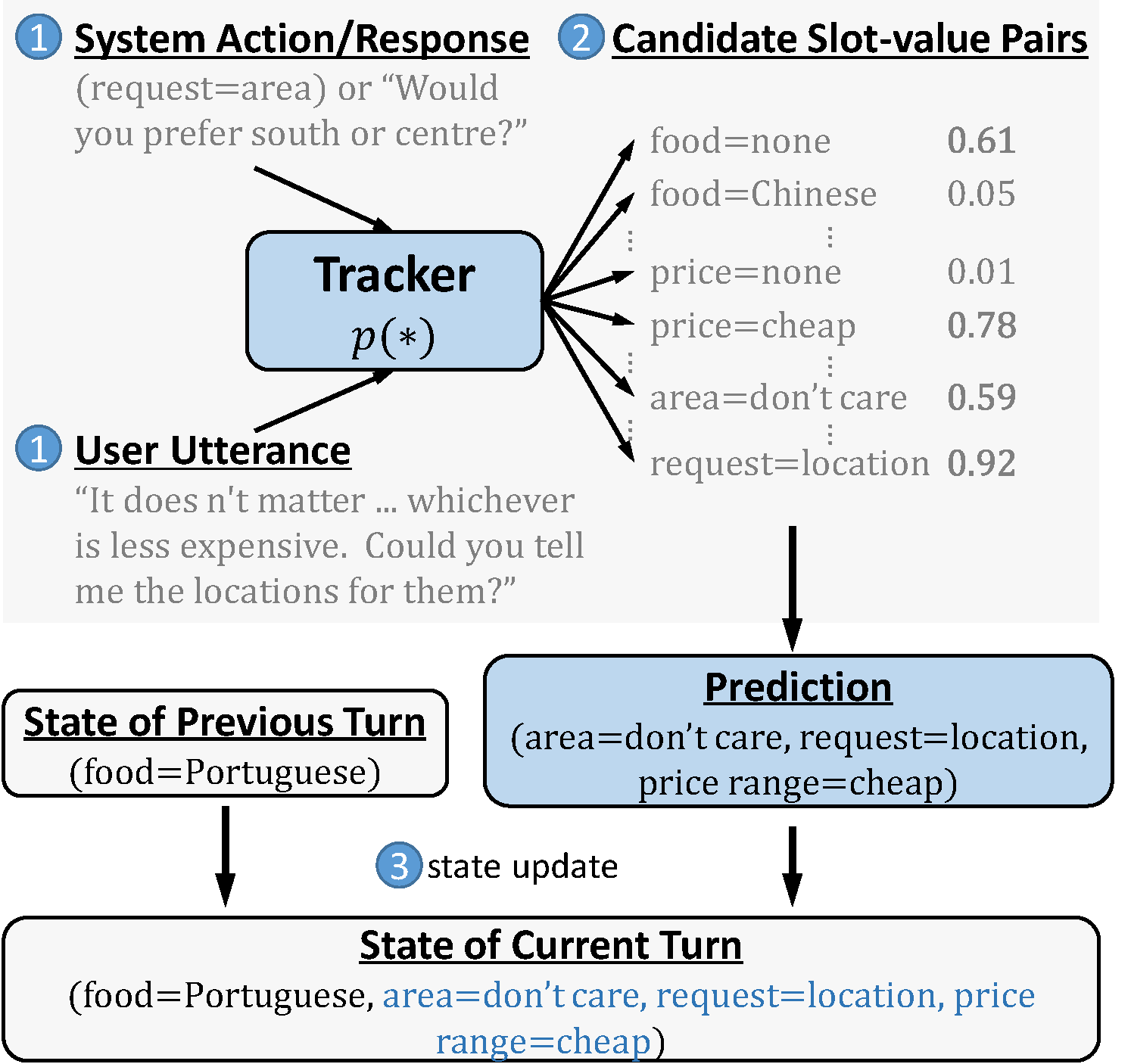}
\caption{The Tracker module. (1) System action or response, and user utterance as input; (2) The tracker predicts the probabilities of all possible slot-value pairs; (3) The prediction and state of previous turn are used to update the state of the current turn.}
\label{fig:tracker_new}
\end{center}
\end{figure}

The dialog state tracker aims to track the user's goal during the dialog process. At each turn, given the user utterance and the system action/response\footnote{If the system actions do not exist in the dataset, we use the system response as the input.}, the trackers first estimate the probabilities of the candidate slot-value pairs\footnote{For each slot, \textit{none} value is added as one candidate slot-value pair.}, and then the pair with the maximum probability for each slot is chosen as the final prediction. To obtain the dialog state of the current turn, trackers typically use the newly predicted slot-values to update the corresponding values in the state of previous turn. One concrete example of the Tracker module is illustrated in Figure~\ref{fig:tracker_new}. 

Our RDA framework is generic and can be applied to different types of tracker models. To demonstrate its effectiveness, we experiment with two different trackers: the state-of-the-art $\text{GLAD}^{\ast}$ model and the classical NBT-CNN (Neural Belief Tracking - Convolutional Neural Networks) model~\cite{mrkvsic2017neural}. The $\text{GLAD}^{\ast}$ is built based on the recently proposed GLAD (Attentive Dialogue State Tracker)~\cite{zhong2018global} by modifying the parameter sharing and attention mechanisms. Due to limited space, we detail the $\text{GLAD}^{\ast}$ model in the Supplementary material. We use the Tracker to refer to $\text{GLAD}^{\ast}$ and NBT-CNN in the following sections.

\subsection{Generator Module}
We formulate data augmentation as an optimal text span replacement problem in a labeled sentence. Specifically, given the tuple of a sentence $\mathbf{x}$, its label $y$, and the text span $p$ of the sentence, the Generator aims to generate a new training instance $(\mathbf{x}',y')$ by substituting $p$ in $\mathbf{x}$ with an optimal candidate $p'$ from a set of candidates for $p$, which we denote as $\mathbf{C}_p$. 

In the span-based data augmentation, we can replace the text span with its paraphrases derived either from existing paraphrase databases or neural paraphrase generation models e.g.~\cite{zhao2009application,D18-1421}. However, directly applying the coarse-grained approach can introduce ineffective or noisy instances for training, and eventually hurt the performance of trackers. Therefore, we train the {Generator} to learn fine-grained generation policies to further improve the quality of the augmented data.  

\noindent\textbf{Generation Process.} The problem of high quality data generation is modeled as a {\it contextual bandit} (or {\it one-step reinforcement learning})~\cite{dudik2011efficient}. Formally, at each trial of a contextual bandit, the context including the sentence $x$ and its text span $p$, is sampled and shown to the agent, then the agent selects a candidate $p'$ from $\mathbf{C}_{p}$ to generate a new instance $\mathbf{x}'$ by replacing $p$ with $p'$.


\vspace{2mm}\noindent\textbf{Policy Learning.} The policy $\pi_\theta(\mathbf{s}, p')$ represents a probability distribution over the valid actions at the current trial, where the state vector $\mathbf{s}$ is extracted from the sentence $\mathbf{x}$, the text span $p$ and the candidate $p'$. $\mathbf{C}_{p}$ forms the action space of the agent given the state $\mathbf{s}$, and the reward $R$ is a scalar value function. The policy is learned to maximize the expected rewards: 
\begin{equation}
\mathcal{J}(\theta)= E_{\pi_{\theta}}[R],
\label{eqn:object}
\end{equation}
where the expectation is taken over state $\mathbf{s}$ and action $p'$.

\begin{figure*}[t]
\begin{center}
\includegraphics[width=15.5cm]{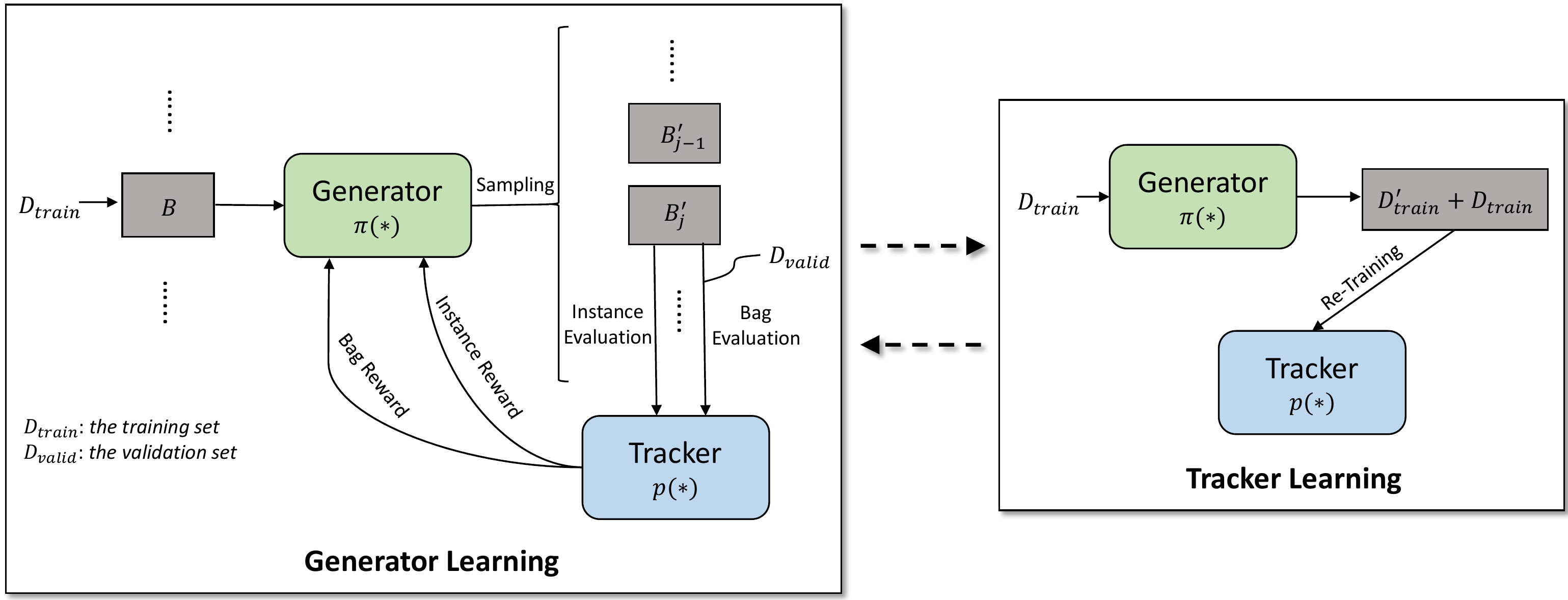}
\caption{The algorithm flow of the reinforced data augmentation framework. The left is the Generator learning and the right is the Tracker learning. The two learning processes are performed in an alternate manner.}
\label{fig:rda_framework_new}
\end{center}
\end{figure*}

The policy $\pi_\theta(\mathbf{s}, p')$ decides which $p' \in \mathbf{C}_p$ to take based on the state $\mathbf{s}$, which is formulated as:
\vspace{-7.5pt}
\begin{equation}
\mathbf{s} = [\mathbf{p}, \mathbf{p}'_{emb}, \mathbf{p}'_{emb} - \mathbf{p}_{emb}, \mathbf{p}'_{emb} \circ \mathbf{p}_{emb}],
\label{eq:state_rep}
\end{equation}
where $\mathbf{p}$ is the contextual representation of $p$, which is derived from the hidden states in the encoder of the {Tracker}, $\mathbf{p}_{emb}$ and $\mathbf{p}'_{emb}$ are the word embeddings of $p$ and $p'$ respectively. For multi-word phrases, we use the average representations of words as the phrase representation. We use a {\it two-layer fully connected network} and {\it sigmoid} to compute the score function $f(\mathbf{s}, p')$ of $p$ being replaced by $p'$. As each $p$ has multiple choices of replacement $\mathbf{C}_p$, we normalize the scores and obtain the final probabilities for the alternative phrases:
\begin{equation}
\pi_\theta(\mathbf{s}, p') = \frac{f(\mathbf{s}, p')}{\sum_{\tilde{p}\in \mathbf{C}_{p}} f(\mathbf{s}, \tilde{p})}.
\end{equation}

The sampling-based \textit{policy gradient} is used to approximate the gradient of the expected reward. To obtain more feedback and make the policy learning more stable, as illustrated in Figure~\ref{fig:rda_framework_new}, we propose to use a two-step sampling method: at first, sample a bag of sentences $B=\{(\mathbf{x}_i, y_i, p_i)\}_{1 \leq i \leq T }$, then iteratively sample a candidate ${p'}_{i,j}$ for each instance in $B$ according to the current policy, obtaining a new bag of instances ${B'}_{j}=\{(\mathbf{x'}_{i,j}, {y'}_{i,j}, {p'}_{i,j})\}_{1 \leq i \leq T }$. After running the bag-level sampling $M$ times, the gradient of objective function can be estimated as:
\vspace{-7.5pt}
\begin{equation}
\scalebox{1}{$
\nabla\mathcal{J}(\theta) \approx \frac{1}{M} \sum_{j=1}^{M}\sum_{i=1}^{T}\nabla_{\theta} \log \pi_\theta(\mathbf{s}_{i,j}, {p'}_{i,j})R_{i,j},$}
\label{eq:gradient}
\end{equation}
where $\mathbf{s}_{i,j}$ and ${p'}_{i,j}$ denote the state and action of the $i$-th instance-level sampling from the $j$-th bag-level sampling, respectively. $R_{i,j}$ is the corresponding reward.

\vspace{2mm}\noindent\textbf{Reward Design.} 
One key problem is assigning suitable rewards to various actions ${p'}_{i,j}$ given state $\mathbf{s}_{i,j}$. 
We design two kinds of rewards: bag-level reward $R^{\text B}_j$ and instance-level reward $R^{\text I}_{i,j}$ in reinforcement learning. The bag-level reward~\cite{feng2018reinforcement,qin2018robust} indicates whether the new sampled bag is helpful to improve the Tracker and the instances in the same bag receive the same reward value. While the instance-level reward assigns different reward values to the instances in the same sampled bags by checking whether the instance can cause the Tracker to make incorrect prediction~\cite{kang18acl,ribeiro-etal-2018-semantically}. We sum two kinds of rewards as the final reward: $R_{i,j}=R^{\text B}_j + R^{\text I}_{i,j}$, for more reliable policy learning.

{\bf Bag-level reward} $R_{j}^B$: we re-train the {Tracker} with each sampled bag and use their performance (e.g., joint goal accuracy~\cite{henderson2014second}) on the validation set to indicate their rewards. Suppose the performance of the $j$-th bag ${B'}_{j}$ is denoted as ${U'}_j$, the bag-level rewards are formulated as:

\vspace{-10pt}
\begin{equation}
R_{j}^{\text B} = \frac{2({U'}_{j} - \min(\{{U'}_{j^*}\}))}{\max(\{{U'}_{j^*}\}) - \min(\{{U'}_{j^*}\})} - 1,
\label{eq:bag_reward}
\end{equation}
where $\{{U'}_{j^*}\}$ refers to the set $\{{U'}_{j^*}\}_{1 \leq j^* \leq M}$. Here we scale the value to be bounded in the range of [-1, 1] to alleviate the instability in RL training\footnote{In this work, the original text span $p$ is also used as one candidate in $\mathbf{C}_p$, which actually acts as an \textit{implicit Baseline}~\cite{sutton2018reinforcement} in RL training.}.

{\bf Instance-level reward} $R_{i,j}^{\text I}$: we evaluate each generated instance $(\mathbf{x}'_{i,j},y'_{i,j})$ in the bag and denote the instance which causes the {Tracker} to make wrong prediction, as {\it large-loss instance} (LI)~\cite{han2018co}. Compared to the non-LIs, the LIs are more informative and can induce larger loss for training the {Tracker}. Thus, in the design of instance-level rewards, the LI is encouraged more when its corresponding bag reward is positive, and punished more when its bag reward is negative. Specifically, we define the instance-level reward as follow:
\vspace{-7.5pt}

\begin{equation}
\scalebox{0.91}{$
  R_{i,j}^{\text I} = \\
 \begin{cases}
{\rm c,} & {R_j^{\text B} \geq 0 \wedge {\mathbb I}_{\text {LI}}(\mathbf{x}'_{i,j},y'_{i,j})}=\text{1}\\
{\rm  {c}/{2},} & {R_j^{\text B} \geq 0 \wedge {\mathbb I}_{\text {LI}}(\mathbf{x}'_{i,j},y'_{i,j})}=\text{0}\\
{\rm  {-c},} & {R_j^{\text B} < 0 \wedge {\mathbb I}_{\text {LI}}(\mathbf{x}'_{i,j},y'_{i,j})}=\text{1}\\
{\rm {-c}/2,} & {R_j^{\text B} < 0 \wedge {\mathbb I}_{\text {LI}}(\mathbf{x}'_{i,j},y'_{i,j})}=\text{0},
 \end{cases}$}
\label{eq:instance_reward}
\end{equation}
where ${\mathbb I}_{\text {LI}}(\mathbf{x}'_{i,j},y'_{i,j})$ is an indicator function of being a LI. We obtain the ${\mathbb I}_{\text {LI}}(\mathbf{x}'_{i,j},y'_{i,j})$ value by checking if the pre-trained Tracker can correctly predict the label on the generated example. ${\rm c}$ is a hyper-parameter, which is set to 0.5 by running a grid search over the validation set.

\renewcommand{\algorithmicrequire}{\textbf{Input:}} 
\renewcommand{\algorithmicensure}{\textbf{Output:}} 

 \begin{algorithm}[t] 
  \caption{The Reinforced Data Augmentation} 
	 \label{alg:RAD_alg}
  \begin{algorithmic}[1] 
   \Require Pre-trained {Tracker} with parameters $\mathnormal{\theta_r}$; the randomly initialized {Generator} with parameters $\mathnormal{\theta_{\pi}}$; 
   \Ensure Re-trained {Tracker}
			\State Store $\theta_{\pi}$
    \For{$l = 1 \to L$}
				\State Re-initialize the {Generator} with $\mathnormal{\theta_{\pi}}$
				\For{$n = 1 \to N $}
					\State Re-initialize the {Tracker} with $\mathnormal{\theta_r}$
					\State Sample a bag $B$
					\For{$j = 1 \to M $}
						\State Sample a new bag ${B}'_{j}$
					\EndFor
					\State Compute bag reward with Eq.~\ref{eq:bag_reward}
					\State Compute instance reward with Eq.~\ref{eq:instance_reward}
					\State Update $\theta_{\pi}$ by the gradients in Eq.\ref{eq:gradient}
				\EndFor
				\State Obtain new data ${D}'$ by the {Generator}
				\State Re-train the {Tracker} on $D + {D}' $, update $\mathnormal{\theta_r}$				
			\EndFor
			\State Save the {Tracker} with $\mathnormal{\theta_r}$ which performs best on the validation set among the $L$ epochs
  \end{algorithmic} 
 \end{algorithm}

\subsection{Alternate Learning}
In the framework of {RDA}, the learning of {Generator} and {Tracker} is conducted in an alternate manner, which is detailed in Algorithm~\ref{alg:RAD_alg} and Figure~\ref{fig:rda_framework_new}. 

The text span $p$ to be replaced has different distribution in the training set. To make learning more efficient, we first sample one text span $p$, then sample one sentence $(\mathbf{x}, y)$ from the sentences containing $p$. This process is made iteratively to obtain a bag $B$. To learn the {Generator}, we generate $M$ bags of instances by running the policy, compute their rewards and update the policy network via the policy gradient method. To learn the {Tracker}, we augment the training data by the updated policy. Particularly for each $(\mathbf{x}, y, p)$, we generate a new instance $(\mathbf{x'}, y', p')$ by sampling based on the learned policies. To further reduce the effect of noisy augmented instances, we remove the new instance if its $p'$ has minimum probability among $\mathbf{C}_p$. We randomly initialize the policy at each epoch to make the generator learn adaptively which policy is best for the current {Tracker}. The alternate learning is performed multiple rounds and the {Tracker} with the best performances on the validation set is saved.

\vspace{-5pt}
\section{Experiment}
\label{experiment}

In this section, we show the experimental results to demonstrate the effectiveness of our framework. 
\vspace{-5pt}
\subsection{Dataset and Evaluation}
We use WoZ~\cite{wen2017network} and MultiWoZ~\cite{budzianowski2018multiwoz} to evaluate the proposed framework on the task of dialog state tracking\footnote{DSTC2~\cite{mrkvsic2017neural} dataset is not used because its clean version~(\url{http://mi.eng.cam.ac.uk/~nm480/dstc2-clean.zip}) is no longer available.}. Following the work~\cite{budzianowski2018multiwoz}, we extract the restaurant domain of the MultiWoZ as the evaluation dataset, denoted as MultiWoZ~(restaurant). Both WoZ and MultiWoZ~(restaurant) are in the restaurant domain. In the experiment, we use the widely used joint goal accuracy~\cite{henderson2014second} as the evaluation metric, which measures  whether all slot values of the updated dialog state exactly match the ground truth values at every turn.

\subsection{Implementation Details}
\label{implement}
We implement the proposed model using PyTorch\footnote{\url{https://pytorch.org/}}. All hyper-parameters of our model are tuned based on the validation set. To demonstrate the robustness of our model, we use the similar hyper-parameter settings for both datasets. Following the previous work~\cite{D18-1299,zhong2018global,nouri2018toward}, we concatenate the pre-trained GloVe embeddings~\cite{pennington2014glove} and the character embeddings~\cite{D17-1206} as the final word embeddings and keep them fixed when training. The epoch number of the alternate learning $L$, the epoch number of the generator learning $N$ and the sampling times $M$ for each bag are set to 5, 200 and 2 respectively.  We set the dimensions of all hidden states to 200 in both the {Tracker} and the {Generator}, and set the head number of multi-head Self-Attention to 4 in the {Tracker}. All learnable parameters are optimized by the ADAM optimizer~\cite{kingma2015adam} with a learning rate of 1e-3. The batch size is set to 16 in the {Tracker} learning, and the bag size in the {Generator} learning is set to 25. 

To avoid over-fitting, we apply dropout to the layer of word embeddings with a rate of 0.2. We also assign rewards based on subsampled validation set with a ratio of 0.3 to avoid over-fitting the policy network on the validation set.

In our experiments, the newly augmented dataset is $n$ times the size of the original training data ($n=5$ for the Woz and $n=3$ for MultiWoz). At each iteration, we randomly sample a subset of the augmented data to train the Tracker. The sampling ratios are 0.4 for Woz and 0.3 for MutiWoz.

For the coarse-grained data augmentation method, we have tried the current neural paraphrase generation model. The preliminary experiment indicates that almost all generated sentences are not helpful for the task of DST. The reason is that most of the neural paraphrase generation models require additional labeled paraphrase corpus which may not be always available~\cite{ray2018robust}. In this work, we extract unigrams, bigrams and trigrams in the training data as the text spans in the generation process. After that, we retrieve the paraphrases for each text span from the PPDB\footnote{\url{http://paraphrase.org/}} database as the candidates. We also use the golden slot value in the sentence as the text spans, the other values of the same slot as the candidates and the label will be changed accordingly.

\begin{table}[tb]
\fontsize{10}{11}\selectfont
\centering
\scalebox{1.0}{
\begin{tabular}{l|c|c}
\toprule
Model  &  WoZ  & Multi\\ 
\midrule
Delexicalised Model & 70.8  & 71.2\\
NBT-DNN & 84.4 & 80.3 \\
NBTKS & 85.5  & 80.9 \\
StateNet & \underline{88.9} & 82.4  \\
GLAD & 88.1 & 82.7\\
GCE & 88.5 & \underline{83.5} \\

\midrule
{NBT-CNN} & $84.0\ {\pm0.6}$ & $79.8\ {\pm 1.0}$ \\
 + {DA} & $84.2\ {\pm0.5}$ & $79.7\ {\pm 0.7} $ \\
 + {RDA} & $87.9^{\ddag}{\pm0.3}$ & $83.4^{\dag}{\pm 0.6}$ \\

\midrule
{$\text{GLAD}^{\star}$} & $88.3\ {\pm 0.3}$ & $83.6\ {\pm 0.9} $ \\
 + {DA} & $88.0\ {\pm0.5}$ & $82.7\ {\pm 0.7}$\\
 + {RDA}&  $\textbf{90.7}^{\ddag}{\pm\textbf{0.2}}$ & $\textbf{86.7}^{\dag}{\pm\textbf{0.5}}$ \\
\bottomrule
\end{tabular}
}
\caption{\label{tab:overall_results} Comparison of our model and other baselines. DA refers the coarse-grained data augmentation without the reinforced framework, and Multi refers the dataset MultiWoZ (restaurant). t-test is conducted in our proposed models and original trackers (NBT-CNN and $\text{GLAD}^{\star}$) are used as the comparison baselines. \dag ~and \ddag: significant over the baseline trackers at 0.05/0.01. The mean and the standard deviation are also reported.}
\end{table}

\subsection{Baseline Methods}
We compare our model with some baselines. \textbf{Delexicalised Model} uses generic tags to replace the slot values and employs a CNN for turn-level feature extraction and a Jordan RNN for state updates~\cite{henderson2014word,wen2017network}.
\textbf{NBT-DNN} and \textbf{NBT-CNN} respectively use the summation and convolution filters to learn the representations for the user utterance, candidate slot-value pair and the system actions~\cite{mrkvsic2017neural}. Then, they fuse these representations by a gating mechanism for the final prediction. 
\textbf{NBTKS} has a similar structure to NBT-DNN and NBT-CNN, but with a more complicated gating mechanism~\cite{ramadan2018large}. 
\textbf{StateNet} learns a representation from the dialog history, and then compares the distances between the learned representation and the vectors of the candidate slot-value pairs for the final prediction~\cite{D18-1299}.
\textbf{GLAD} is a global-locally self-attentive state tracker, which learns representations of the user utterance and previous system actions with global-local modules~\cite{zhong2018global}.
\textbf{GCE} is developed based on GLAD by using global recurrent networks rather than the global-local modules~\cite{nouri2018toward}. 

We also use the coarse-grained data augmentation (\textbf{DA}) without the reinforced framework as the baseline, which generate new instances by randomly choosing one from the candidates.

\begin{table}[tb]
\fontsize{10}{11}\selectfont
\centering
\scalebox{1.0}{
\begin{tabular}{l|l|ccc}
\toprule
Dataset & Model & 10\% & 20\% & 50\% \\ 
\midrule
\multirow{2}{*}{WoZ}& {$\text{GLAD}^{\star}$} & 50.1 & 72.5 & 81.7  \\

& + {RDA} & \textbf{66.8} & \textbf{81.5} & \textbf{86.9} \\
\midrule
\multirow{2}{*}{Multi}& {$\text{GLAD}^{\star}$} & 60.0 &  72.6  & 77.6  \\

& + {RDA} & $\textbf{71.5}$ & $\textbf{81.2}$ & $\textbf{85.2}$ \\
\bottomrule
\end{tabular}
}
\caption{\label{tab:subsampling_results} The results with different sub-sampling ratios on WoZ and MultiWoZ~(restaurant). }
\end{table}

\begin{table}[tb]
\fontsize{10}{11}\selectfont
\centering
\scalebox{1.0}{
\begin{tabular}{l|c|c}
\toprule
Setting  &  WoZ  & Multi\\ 
\midrule
RDA & \textbf{90.7} & \textbf{86.7} \\
- Bag Reward & 89.1 & 84.3 \\
- Instance Reward & 89.8 & 85.4 \\
\midrule
DA & 88.0 & 82.7 \\
\bottomrule
\end{tabular}
}
\caption{\label{tab:ablation_results} Ablation study of performances on the test set of WoZ and MultiWoZ.}
\end{table}

\subsection{Results and Analyses} 
We compare our model with baselines and the joint goal accuracy is used as the evaluation metric. The results are shown in Table~\ref{tab:overall_results}.

From the table, we observe that the proposed~{$\text{GLAD}^{\star}$} achieves comparable performances (88.3\% and 83.6\%) with other state-of-the-art models on both datasets. The framework {RDA} can further boost the performances of the competitive~{$\text{GLAD}^{\star}$} by the margin of 2.4\% and 3.1\% on two datasets respectively, achieving new state-of-the-art results (90.7\% and 86.7\%).  Compared with the $\text{GLAD}^{\star}$, the classical NBT-CNN with the RDA framework obtains more improvements: 3.9\% and 3.6\%. We also conduct significance test (t-test), and the results show that the proposed RDA achieves significant improvements over baseline models ($p<0.01$ and $p<0.05$ respectively for WoZ and MultiWoZ~(restaurant)).

The table also shows that directly using coarse-grained data augmentation methods without the~{RDA} is less effective, and can even degrade the performances, as it may generate noisy instances. The results show that: using the RDA, the $\text{GLAD}^{\star}$ achieves improvements of (88.0\%$\rightarrow$90.7\%) and (82.7\%$\rightarrow$86.7\%) respectively on the WoZ and MultiWoZ. The NBT-CNN obtains improvements of (84.2\%$\rightarrow$87.9\%) and (79.7\%$\rightarrow$83.4\%) respectively. Overall, the results indicate that the {RDA} framework offers an effective mechanism to improve the quality of augmented data.

To further verify the effectiveness of the {RDA} when the training data is scarce, we conduct sub-sampling experiments with the $\text{GLAD}^{\star}$ tracker trained on different ratios [10\%, 20\%, 50\%] of the training set. The results on both datasets are shown in Table~\ref{tab:subsampling_results}. We find that our proposed {RDA} methods consistently improve the original~{tracker} performance. Notably, we obtain $\sim$10\% improvements with [10\%, 20\%] ratios of training set on both WoZ and MultiWoZ~(restaurant), which indicates that the RDA framework is particularly useful when the training data is limited.

To evaluate the performance of different level rewards, we perform ablation study with {$\text{GLAD}^{\star}$} on both the WoZ and MultiWoz datasets. The results are shown in Table~\ref{tab:ablation_results}. From the table we can see that both rewards can provide the improvements of 1\% to 2\% in the datasets and the bag-level reward achieves larger gains than the instance-level reward. Compared with DA setting, RDA obtains the improvements of 3\% to 4\% on the datasets by combining the both rewards, which indicates that the summation reward is more reliable for policy learning than individual ones. 
\subsection{Effects of Hyper-parameters}
\begin{figure}[ht]

\begin{center}
\includegraphics[width=0.85\linewidth]{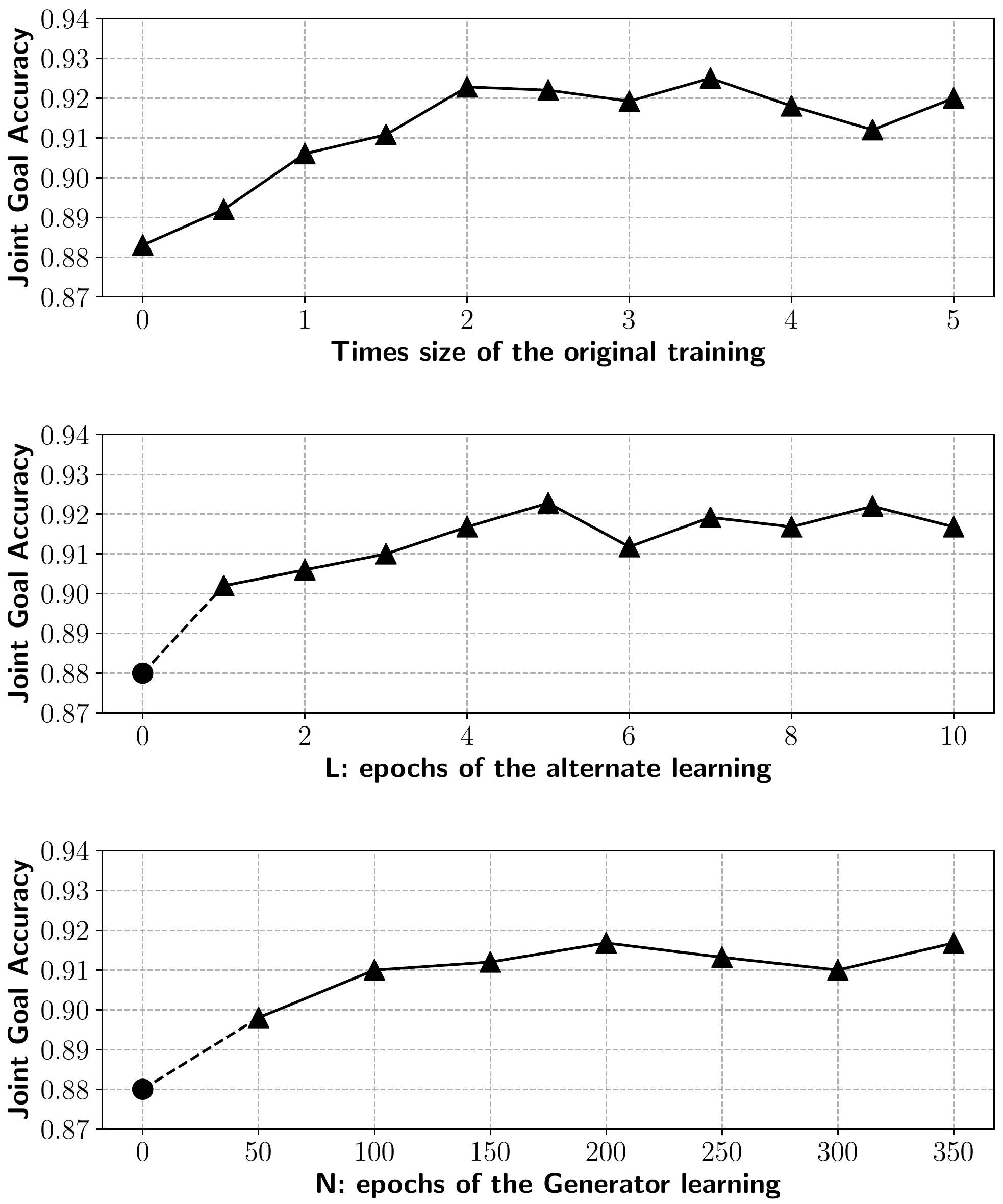}
\caption{Results of different hyper-parameters. Top: different times the size of original data; Middle: different epochs of alternate learning; Bottom: different epochs of the Generator learning. The solid circles of $L = 0$ and $N = 0$ in the figure refer to the model of coarse-grained data augmentation (DA).}
\label{fig:hyper_parameters}
\end{center}
\end{figure}
\noindent In this subsection, we investigate the effects of the number of newly augmented data in the Tracker learning, the epoch number of the alternate learning $L$ and the epoch number of the Generator learning $N$ on performance. We conduct experiments with the $\text{GLAD}^{\star}$ tracker which is evaluated on the validation set of WoZ and the joint goal accuracy is used as the evaluation metric.

\noindent \textbf{Number of newly augmented data}: we use 0 to 5 times the size of original data in the Tracker learning. The performance is shown in Figure~\ref{fig:hyper_parameters}~(top). The model continues to improve when the number of newly added examples is less than 2 times the original data. When we add more than twice the amount of original data, the improvement is not significant. \\
\textbf{Epoch number of the alternate learning}: we vary $L$ from 0 to 10 and the performance is shown in Figure~\ref{fig:hyper_parameters} (middle). We can see that, with alternate learning, the model continues to improve when $L \leq 5$, and becomes stable with no improvement after $L > 5$. \\
\textbf{Epoch number of the Generator learning}: we vary $N$ from 0 to 350, and the performance is shown in Figure~\ref{fig:hyper_parameters} (bottom). We find that the performance increases dramatically when $N\leq 200$, and shows no improvement after $N > 200$. It shows that the Generator needs a large $N$ to ensure a good policy.

\vspace{-2mm}
\subsection{Case Study for Policy Network}
\noindent We sample four sentences from WoZ to demonstrate the effectiveness of the Generator policy in the case study. Due to limited space, we present the candidate phrases with maximum and minimum probabilities derived from the policy network and the details are shown in Table~\ref{tab:case_study}.

We observe that both high-quality and low-quality replacements exist in the candidate set. The high-quality replacements will generate reliable instances, which can potentially improve the generalization ability of the Tracker. The low-quality ones will induce noisy instances and can reduce the performance of the Tracker. From the results of the policy network, we find that our Generator can automatically infer the quality of candidate replacements, assigning higher probabilities to the high-quality candidates and lower probabilities to the low-quality candidates.

\newcommand{\tabincell}[2]{\begin{tabular}{@{}#1@{}}#2\end{tabular}} 
\begin{table}[h]
\fontsize{10}{11}\selectfont
\centering
\scalebox{0.735}{
\begin{tabular}{l|l}
\toprule
Sentence $\mathbf{x}$ and text span $p$ & Candidates $\mathbf{C}_p$  \\ 
\midrule
\multirow{2}{*}{\tabincell{l}{Thanks , [\textbf{could you give}] me the \\ phone number for the restaurant?}}& {i was wonder if you could provide} \\
& {are you able to} \\
\midrule
\multirow{2}{*}{\tabincell{l}{What restaurants are on the east \\ side that are not [\textbf{overpriced}] ?}} &{too expensive} \\
&{cheap enough}  \\
\midrule
\multirow{2}{*}{\tabincell{l}{What is a affordable restaurant in \\ the [\textbf{south side part}] of town?}} & {south end} \\
& {southern countries}  \\
\midrule
\multirow{2}{*}{\tabincell{l}{I want Cuban food and i [\textbf{do n't} \\ \textbf{care}] about the price range. }} &{do n't worry}  \\
& {do n't give a danm}  \\
\bottomrule
\end{tabular}
}
\caption{\label{tab:case_study} Case study for the Generator policy. The phrases with maximum policy values are listed at the first line in each cell of Candidates $\mathbf{C}_p$ and the ones with minimum values are listed at the second line.}
\end{table}


\section{Related Work}
\label{related_work}
\textbf{Dialog State Tracking.}
DST is studied extensively in the literature~\cite{williams2016dialog}. The methods can be classified into three categories: rule-based~\cite{zue2000juplter}, generative~\cite{devault2007managing,williams2008exploiting}, and discriminative ~\cite{metallinou2013discriminative} methods. The discriminative methods~\cite{metallinou2013discriminative} study dialog state tracking as a classification problem, designing a large number of features and optimizing the model parameters by the annotated data. Recently, neural networks based models with different architectures have been applied in DST~\cite{henderson2014word,zhong2018global}. These models initially employ CNN~\cite{wen2017network}, RNN~\cite{ramadan2018large}, self-attention~\cite{nouri2018toward} to learn the representations for the user utterance and the system actions/response, then various gating mechanisms~\cite{ramadan2018large} are used to fuse the learned representations for prediction. Another difference among these neural models is the way of parameter sharing, most of which use one shared global encoder for representation learning, while the work~\cite{zhong2018global} pairs each slot with a local encoder in addition to one shared global encoder.
Although these neural network based trackers obtain state-of-the-art results, they are still limited by insufficient amount and diversity of annotated data. To address this difficulty, we propose a method of data augmentation to improve neural state trackers by adding high-quality generated instances as new training data.

\noindent\textbf{Data Augmentation.} 
Data augmentation aims to generate new training data by conducting transformations (e.g. rotating or flipping images, audio perturbation, etc.) on existing data. It has been widely used in computer vision~\cite{krizhevsky2012imagenet,cubuk2018autoaugment} and speech recognition~\cite{ko2015audio}. In contrast to image or speech transformations, it is difficult to obtain effective transformation rules for text which can preserve the fluency and coherence of newly generated text and be useful for specific tasks. There is prior work on data augmentation in NLP~\cite{zhang2015character,kang18acl,kobayashi2018contextual,hou2018sequence,ray2018robust,yoo2018data}. These approaches do not specially design some mechanisms to filter out low-quality generated instances. In contrast, we propose a coarse-to-fine strategy for data augmentation, where the fine-grained generative polices learned by RL are used to automatically reduce the noisy instances and retain the effective ones. 

\noindent\textbf{Reinforcement Learning in NLP.} RL is a general purpose framework for decision making and has been applied in many NLP tasks such as information extraction~\cite{narasimhan2016improving}, relational reasoning~\cite{xiong2017deeppath}, sequence learning~\cite{ranzato2015sequence,D18-1421,celikyilmaz2018deep}, summarization~\cite{paulus2017deep,dong2018banditsum}, text classification~\cite{wu2018reinforced,feng2018reinforcement} and dialog~\cite{singh2000reinforcement,D16-1127}. Previous works by~\cite{feng2018reinforcement} and~\cite{P18-1046} design RL algorithm to learn how to filter out noisy ones. Our work is significantly different from these works, especially in the problem settings and model frameworks. The previous work assume there are many distant sentences. However, in our work we only know possible replacements, and our RL algorithm should learn how to choose optimal replacements to “generate” new high-quality sentences. Moreover, the action space and reward design are different.

\section{Conclusion and Future Work}
\label{conclusion}
We have proposed a reinforced data augmentation (RDA) method for dialogue state tracking in order to improve its performance by generating high-quality training data. The Generator and the Tracker are learned in an alternate manner, i.e. the Generator is learned based on rewards from the Tracker while the Tracker is re-trained and boosted with the new high-quality data augmented by the Generator. We conducted extensive experiments on the datasets of WoZ and MultiWoZ~(restaurant); the results demonstrate the effectiveness of our framework. In future work, we would conduct experiments on more NLP tasks and introduce neural network based paraphrasing method in the RDA framework.

\newpage

\bibliography{emnlp}
\bibliographystyle{acl_natbib}

\end{document}